\documentclass[11pt]{article}

\usepackage[final]{acl}

\usepackage{times}
\usepackage{latexsym}
\usepackage{amsmath}
\usepackage{booktabs}
\usepackage{multirow}
\usepackage[table]{xcolor}
\usepackage{xcolor}
\usepackage{tabularx}
\usepackage[normalem]{ulem}
\usepackage{float}
\usepackage{enumitem}
\usepackage{setspace}

\newcommand{\posdelta}[1]{%
  {\scriptsize(#1)}%
}

\newcommand{\smalldrop}[1]{%
  {\scriptsize(#1)}%
}

\newcommand{\bigdrop}[1]{%
  {\textcolor{red!75!black}
  {\textbf{\scriptsize(#1)}}}%
}

\newcommand{\biggain}[1]{%
  {\textcolor{green!60!black}
  {\textbf{\scriptsize(#1)}}}%
}

\usepackage{tcolorbox}
\tcbuselibrary{skins, breakable}
\newtcolorbox{promptbox}[1]{
  colback=gray!15!white, 
  colframe=gray!85!black, 
  title=\textbf{#1}, 
  fonttitle=\bfseries\sffamily,
  enhanced,
  boxrule=0.8pt,
  left=5pt, right=5pt, top=5pt, bottom=5pt
}

\usepackage[T1]{fontenc}

\usepackage[utf8]{inputenc}

\usepackage{microtype}

\usepackage{inconsolata}

\usepackage{graphicx}

\title{DeCode: Decoupling Content and Delivery for Medical QA}

\author{
Po-Jen Ko\thanks{Work done during internship at HTC DeepQ} \\
National Taiwan University \\
\texttt{b11902138@ntu.edu.tw}
\And
Chen-Han Tsai \\
HTC DeepQ \\
\texttt{maxwell\_tsai@htc.com}
\And
Yu-Shao Peng \\
HTC DeepQ \\
\texttt{ys\_peng@htc.com}
}

\begin{document}
\maketitle

\begin{abstract}
Large language models (LLMs) exhibit strong medical knowledge and can generate factually accurate responses. However, existing models often fail to account for individual patient contexts, producing answers that are clinically correct yet poorly aligned with patients’ needs. In this work, we introduce \textbf{DeCode} (\textbf{De}coupling \textbf{Co}ntent and \textbf{De}livery), a training-free, model-agnostic framework that adapts existing LLMs to produce contextualized answers in clinical settings. We evaluate DeCode on OpenAI HealthBench, a comprehensive and challenging benchmark designed to assess clinical relevance and validity of LLM responses. DeCode boosts zero-shot performance from $28.4\%$ to $49.8\%$ and achieves new state-of-the-art compared to existing methods.
Experimental results suggest the effectiveness of DeCode in improving clinical question answering of LLMs.


\end{abstract} 


\section{Introduction}
Large language models (LLMs) have recently achieved strong performance on a variety of medical natural language processing tasks, most notably medical question answering (QA), where models are evaluated on their ability to generate correct responses to clinically relevant questions \citep{singhal2025toward, nori2023can}. This progress has been demonstrated across a growing collection of medical QA benchmarks, spanning multiple-choice and generative settings, professional examination-style questions, and open-domain clinical knowledge assessments \citep{jin2021disease, pal2022medmcqa}. Collectively, results on these evaluations suggest that contemporary LLMs exhibit substantial medical knowledge and reasoning capability under standardized testing conditions \citep{saab2024capabilities, openai2024o1}.

Existing medical QA benchmarks, however, are predominantly designed to measure answer correctness or reasoning accuracy, often via exact-match, multiple-choice selection, or expert-graded factual validity. While these metrics are well-suited for assessing knowledge recall and clinical reasoning, they provide only a partial characterization of model behavior in patient-facing or clinical communication settings \citep{gong2025knowledge, tu2025conversational}. In particular, such evaluations do not capture whether model responses are understandable, appropriately calibrated to patient context, or aligned with norms of safe and empathetic medical communication.

This limitation motivates the need for evaluation frameworks that extend beyond accuracy-based metrics. OpenAI HealthBench was introduced to address this gap by evaluating medical LLM outputs along multiple qualitative dimensions, including context seeking, emergency referrals, and responding under uncertainty, in addition to factual correctness \citep{arora2025healthbench}. Unlike prior medical QA datasets, which typically assume a single correct answer independent of delivery style or audience, HealthBench explicitly models the interactional aspects of medical responses, enabling a more fine-grained analysis of clinically relevant response quality.

Empirical results on HealthBench further show that models with comparable accuracy on traditional medical QA benchmarks can exhibit substantial variation across other non-accuracy dimensions, revealing a misalignment between standardized QA performance and patient-centered context awareness \citep{arora2025healthbench}. Together, these findings suggest that accuracy alone is insufficient as a proxy for real-world clinical readiness and underscore the importance of multidimensional evaluation for medical LLMs.

In this work, we introduce the \textbf{Decoupling Content and Delivery (DeCode)} framework, a modular approach for generating patient-specific medical responses from clinical conversations. DeCode decomposes an existing clinical interaction into multiple complementary analytical perspectives, each implemented via a specialized LLM module. The outputs of these modules are subsequently synthesized to produce a final response that accounts for both medical correctness and patient context.

Importantly, DeCode operates in a training-free paradigm, orchestrating the generation process through explicit clinical formulation and structured discourse constraints. 
Empirically, we demonstrate that DeCode substantially improves performance on HealthBench, boosting zero-shot performance from $28.4\%$ to $49.8\%$ and surpassing the recent state-of-the-art, MuSeR ($47.1\%$), by an absolute margin of $2.7\%$.
Furthermore, we show that DeCode generalizes consistently across multiple leading LLMs, suggesting that the framework captures model-agnostic principles for personalized medical response generation.

The remainder of this paper is organized as follows. Related works are introduced in Section~\ref{sec:related}. The proposed method is presented in Section~\ref{sec:method}. Experimental setup and results are provided in Section~\ref{sec:setup} and Section~\ref{sec:results}, respectively. Finally, Section~\ref{sec:conclusion} concludes the paper.

\section{Related Work}
\label{sec:related}
Early evaluations of large language models (LLMs) in medical question answering have primarily focused on standardized multiple-choice benchmarks, including MedQA~\cite{jin2021disease}, MedMCQA~\cite{pal2022medmcqa}, and PubMedQA~\cite{jin2019pubmedqa}. These benchmarks have catalyzed substantial research on assessing and improving medical knowledge in LLMs~\cite{singhal2025toward, nori2023can, saab2024capabilities, jeong2024improving, li2024agenthospital, wu2025automedprompt}. However, such evaluations remain inherently static and accuracy-centric, limiting their ability to assess communicative competence, contextual sensitivity, and patient-centered delivery beyond factual correctness~\cite{gong2025knowledge}.

HealthBench~\cite{arora2025healthbench} introduces a multidimensional evaluation framework for medical QA based on open-ended, multi-turn clinical conversations. Unlike traditional multiple-choice benchmarks, HealthBench employs physician-authored rubrics to assess behavioral dimensions such as clinical accuracy, communication quality, and contextual awareness, enabling a more comprehensive evaluation of medical QA systems beyond factual correctness.

MuSeR~\cite{zhou2025muser} targets HealthBench by proposing a self-refinement framework in which a student LLM is guided by high-quality responses from a reference teacher model. In its original formulation, MuSeR relies on data synthesis and supervised training: the student generates an initial response, performs structured self-assessment across multiple dimensions, and produces a refined final answer. While this training-based pipeline is computationally intensive and primarily applicable to trainable, open-source LLMs, the core self-refinement procedure can also be applied at inference time, enabling response refinement without model distillation or fine-tuning.

In parallel, multi-agent frameworks have been proposed to address complex medical QA by decomposing reasoning across specialized roles. MedAgents~\cite{tang2024medagents} employs role-playing specialists for debate-based hypothesis refinement, while MDAgents~\cite{kim2024mdagents} dynamically configures expert teams based on query complexity. More recent approaches further extend this paradigm: KAMAC~\cite{wu2025kamac} introduces on-demand expert recruitment to address knowledge gaps during generation, and AI Hospital~\cite{fan2025aihospital} evaluates agent-based systems in interactive patient simulation environments. However, these methods primarily emphasize diagnostic reasoning and accuracy on traditional benchmarks, often overlooking how complex reasoning outcomes are translated into clear, user-aligned responses.



Building on these observations, we introduce DeCode, a modular framework that explicitly decouples medical content reasoning from response delivery. Unlike training-based or agent-centric approaches, DeCode requires no additional training and is model-agnostic, while emphasizing structured generation that supports contextualized and user-aligned medical responses. We present our implementation in the following section.

\section{Method}
\label{sec:method}
\begin{figure*}[t]
    \centering
    \includegraphics[width=\textwidth]{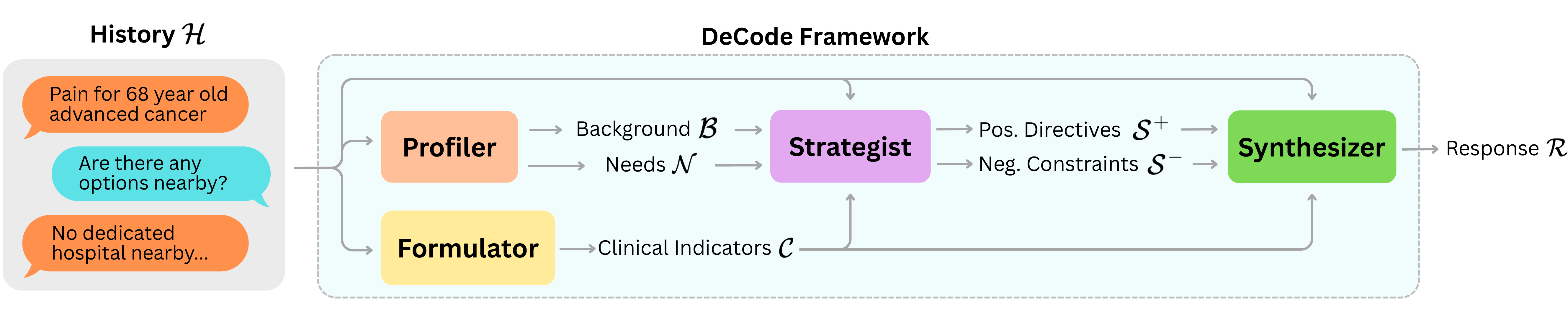}
    \caption{\textbf{The DeCode Framework Pipeline.} Given the conversation history $\mathcal{H}$, the system first employs the \textbf{Profiler} and \textbf{Formulator} to extract user context ($\mathcal{B}, \mathcal{N}$) and clinical indicators $\mathcal{C}$. These components are then synthesized by the \textbf{Strategist} to generate tailored directives $\mathcal{S}$ (consisting of positive strategies $\mathcal{S}^+$ and negative constraints $\mathcal{S}^-$). Finally, the \textbf{Synthesizer} constructs the response based on $\mathcal{C}$ and $\mathcal{S}$, ensuring both medical accuracy and user adaptability.}
    \label{fig:overall_pipeline}
\end{figure*} 


Medical question answering with LLMs can be modeled as a form of conditional text generation $P(\mathcal{R}\mid\mathcal{H})$, where $\mathcal{R}$ denotes the response and $\mathcal{H}$ the conversation history. In practice, $\mathcal{H}$ contains rich patient-specific information---such as symptoms, risk factors, and health indicators---distributed across multiple dialogue turns. However, LLMs are typically trained to model this distribution directly, without mechanisms to explicitly aggregate these dispersed signals. As a result, specific patient details are frequently overlooked during response generation.

To address this limitation, we introduce DeCode, a framework that structures the generation process through four intermediate textual representations: user background $\mathcal{B}$, user needs $\mathcal{N}$, clinical indicators $\mathcal{C}$, and discourse strategy $\mathcal{S}$. As illustrated in Figure 1, these representations are orchestrated by four corresponding modules: Profiler $\mathcal{M}_{prof}$, Formulator $\mathcal{M}_{form}$, Strategist $\mathcal{M}_{strat}$, and Synthesizer $\mathcal{M}_{syn}$. By disentangling \textit{content} from \textit{delivery}, DeCode enables independent optimization of medical accuracy and communicative quality. The inference process is formalized as a sequential chain:
\begin{equation*}
\begin{split}
    \mathcal{R} = \, & \underbrace{\mathcal{M}_{syn}(\mathcal{S}, \mathcal{C}, \mathcal{H})}_{\text{Synthesis}} \circ \underbrace{\mathcal{M}_{strat}(\mathcal{B}, \mathcal{N}, \mathcal{C}, \mathcal{H})}_{\text{Strategy}} \\
    & \circ \underbrace{\{\mathcal{M}_{prof}(\mathcal{H}), \mathcal{M}_{form}(\mathcal{H})\}}_{\text{Extraction}}
\end{split}
\end{equation*}
where $\mathcal{M}$ denotes the LLM modules tailored for specific sub-tasks. In the following sections, we detail the design of each module.

\subsection{Profiler: User Context Disentanglement}
\label{sec:user_profiling}
Medical advice varies significantly across individuals. The same symptom may imply different risks depending on the user's background and lifestyle. 
To capture this nuance beyond surface-level queries, the Profiler $\mathcal{M}_{prof}$ extracts the user's specific context from the conversation history $\mathcal{H}$. We formalize this extraction as:
\begin{equation*}
    (\mathcal{B}, \mathcal{N}) = \mathcal{M}_{prof}(\mathcal{H}).
\end{equation*}
The user background $\mathcal{B}$ encapsulates critical attributes such as age, occupation, and living conditions that constrain actionable advice. Concurrently, the user needs $\mathcal{N}$ identifies the user's core intent by synthesizing the conversation history $\mathcal{H}$. By decoupling the user information $\mathcal{B}$ and $\mathcal{N}$ from the history $\mathcal{H}$, we allow improved clarity in identifying user-specific constraints during response formulation. The user background $\mathcal{B}$ and user needs $\mathcal{N}$ are then sent to the Strategist module.

\subsection{Formulator: Clinical Distillation}
\label{sec:clinical_content}
A critical challenge in medical dialogue is that diagnostic cues are often dispersed throughout the conversation history $\mathcal{H}$, making it difficult to verify if the response covers all relevant medical aspects. To address this, the Formulator $\mathcal{M}_{form}$ functions as a clinical information distiller. It extracts and aggregates a structured set of clinical indicators $\mathcal{C}$ (e.g., symptoms, possible causes, and potential red flags) from the user statements in $\mathcal{H}$. We formalize this process as
\begin{equation*}
    \mathcal{C} = \mathcal{M}_{form}(\mathcal{H}).
\end{equation*}


Crucially, this module operates purely on a factual level, decoupling the medical substance from the delivery style. By explicitly manifesting $\mathcal{C}$ as an intermediate representation, the system provides a rigorous checklist for the downstream modules. This ensures that the final response is grounded in verified medical evidence and that high-stakes safety indicators are duly addressed, regardless of the chosen empathy level or conversation tone.



\subsection{Strategist: Discourse Orchestration}
\label{sec:discourse_strategy}

Beyond factual accuracy, effective medical dialogue requires determining the optimal delivery strategy tailored to the user's cognitive and emotional context. The Strategist $\mathcal{M}_{strat}$ addresses this gap by synthesizing the conversation history $\mathcal{H}$, extracted user profile ($\mathcal{B}, \mathcal{N}$) and clinical indicators ($\mathcal{C}$) into a coherent strategy. We formalize this process as:
\begin{equation*}
    \mathcal{S} = \{ \mathcal{S}^{+}, \mathcal{S}^{-} \} = \mathcal{M}_{strat}(\mathcal{B}, \mathcal{N}, \mathcal{C}, \mathcal{H}).
\end{equation*}

The resulting discourse strategy $\mathcal{S}$ comprises two complementary sets. Positive directives $\mathcal{S}^{+}$ prescribe the prioritization of clinical content and establish the appropriate level of technical detail, crucially instructing the model to actively seek clarification when information is insufficient. Conversely, negative constraints $\mathcal{S}^{-}$ serve as behavioral guardrails, preventing counterproductive styles (e.g., overly academic tones) and filtering out content that may be overwhelming or potentially misleading for the specific user. By enforcing these strategies, the module ensures that the final response is not only medically grounded but also empathetic and strictly aligned with the user's preferences.





\subsection{Synthesizer: Controlled Generation}
\label{sec:response_generation}
Finally, the Synthesizer $\mathcal{M}_{syn}$ generates the response $R$ by integrating the clinical indicators $\mathcal{C}$ with the discourse strategy $\mathcal{S}$. We formalize this process as:
\begin{equation*}
    R = \mathcal{M}_{syn}(\mathcal{S}, \mathcal{C}, \mathcal{H}).
\end{equation*}
By separating content formulation from delivery planning, the Synthesizer operates as a constrained generator. It articulates the verified information in $\mathcal{C}$ while adhering to the directives defined in $\mathcal{S}$. This ensures that the generation process focuses on realization rather than reasoning, producing outputs that are clinically accurate and contextually appropriate.

Taken together, the Profiler, Formulator, Strategist, and Synthesizer form a coherent generation pipeline that transforms the conversation history $\mathcal{H}$ into a personalized and clinically grounded response $R$. Each module addresses a distinct stage of the reasoning–generation process, enabling explicit control over user understanding, clinical content, discourse planning, and surface realization. For reproducibility and clarity, the prompts corresponding to each module are provided in the appendix.


\section{Experiments}
\label{sec:setup}
\subsection{Dataset and Evaluation}
We evaluate on OpenAI HealthBench~\citep{arora2025healthbench}, which contains $5{,}000$ simulated multi-turn patient--clinician conversations ending in a user query. Each conversation is annotated by medical professionals and assigned to one of seven themes: \textit{emergency referrals}, \textit{context seeking}, \textit{global health}, \textit{health data tasks}, \textit{complex responses}, \textit{hedging}, and \textit{communication}. HealthBench additionally provides physician-authored rubrics per conversation, grouped into five evaluation axes: \textit{accuracy}, \textit{completeness}, \textit{communication quality}, \textit{context awareness}, and \textit{instruction following}. For more details regarding the conversation themes and evaluation axes, please refer to the original paper \cite{arora2025healthbench}.

\paragraph{Metric.}
We follow the official HealthBench protocol~\citep{arora2025healthbench}: each conversation is graded using its rubric and scored by GPT-4.1~\cite{openai2025gpt4.1}. Reported numbers are the mean normalized score over the evaluated set.

\paragraph{Splits.}
We report results on the full HealthBench dataset for the primary evaluation. Owing to the computational cost, all subsequent experiments are conducted on the Hard subset of $1{,}000$ challenging conversations.

\subsection{Implementation Details}
\paragraph{Base LLMs.}
We use OpenAI~o3~\cite{openai2025o3} as the primary base model and additionally evaluate GPT-5.2~\cite{openai2025gpt5.2}, Claude-Sonnet~4.5~\cite{anthropic2025claude45}, and DeepSeek~R1~\cite{deepseekr1} to assess generalization across model families.

\paragraph{Comparison Methods.}
We compare against: (i) standard \textbf{Zero-shot} (ii) \textbf{Single Stage}, an improved baseline that compresses the four phases of DeCode into a single prompt; (iii) \textbf{MDAgents}~\citep{kim2024mdagents}, which recruits specialized experts based on query complexity and aggregates their responses; (iv) \textbf{KAMAC}~\citep{wu2025kamac}, which dynamically recruits experts during generation; and (v) \textbf{MuSeR}~\citep{zhou2025muser}, which applies a self-refinement generation strategy. For MuSeR, we use the authors’ self-refinement prompts at inference time, without model distillation or fine-tuning. A detailed cost and latency breakdown for all frameworks is included in the appendix.

\section{Results and Analysis}
\label{sec:results}

\subsection{Main Results}
\label{results:main_results}
\begin{table*}[!t]
\centering
\caption{\textbf{Comparison between zero-shot prompting and DeCode.} Both methods use OpenAI~o3 as the base LLM and are evaluated on the full HealthBench dataset and its hard subset. DeCode consistently outperforms the zero-shot baseline across most conversation themes and evaluation axes. Performance on the hard subset is substantially lower than on the full set, highlighting the increased difficulty of this evaluation setting. Deltas in DeCode columns are computed relative to the zero-shot baseline within the same split; deltas greater than 2 points are highlighted.}
\label{tab:main_results}
\resizebox{0.9\textwidth}{!}{
\renewcommand{\arraystretch}{0.91}
\begin{tabular}{l cc|cc}
\toprule
& \multicolumn{2}{c|}{\textbf{Full Set}} & \multicolumn{2}{c}{\textbf{Hard Subset}} \\
\cmidrule(lr){2-3} \cmidrule(lr){4-5}
\textbf{Metric}~($\uparrow$, \%) & \textbf{Zero-shot} & \textbf{DeCode (Ours)} & \textbf{Zero-shot} & \textbf{DeCode (Ours)} \\
\midrule
\rowcolor{gray!10} \textbf{Overall Score} 
& 57.8 & \textbf{67.8} \biggain{+10.0}
& 28.4 & \textbf{49.8} \biggain{+21.4} \\
\midrule
\multicolumn{5}{l}{\textit{\textbf{Themes}}} \\
Emergency Referrals 
& 69.2 & \textbf{80.3} \biggain{+11.1}
& 27.0 & \textbf{59.1} \biggain{+32.1} \\
Context Seeking 
& 51.2 & \textbf{67.0} \biggain{+15.8}
& 30.0 & \textbf{58.3} \biggain{+28.3} \\
Global Health 
& 52.7 & \textbf{65.2} \biggain{+12.5}
& 31.8 & \textbf{49.8} \biggain{+18.0} \\
Health Data Tasks 
& 44.3 & \textbf{56.7} \biggain{+12.4}
& 17.0 & \textbf{35.6} \biggain{+18.6} \\
Communications 
& 67.9 & \textbf{74.4} \biggain{+6.5}
& 29.2 & \textbf{43.8} \biggain{+14.6} \\
Hedging 
& 59.6 & \textbf{69.5} \biggain{+9.9}
& 30.9 & \textbf{54.6} \biggain{+23.7} \\
Complex Responses 
& \textbf{55.1} & 52.6 \bigdrop{-2.5}
& 24.4 & \textbf{41.3} \biggain{+16.9} \\
\midrule
\multicolumn{5}{l}{\textit{\textbf{Axes}}} \\
Accuracy 
& 66.4 & \textbf{72.5} \biggain{+6.1}
& 45.6 & \textbf{54.3} \biggain{+8.7} \\
Completeness 
& 59.5 & \textbf{74.0} \biggain{+14.5}
& 30.7 & \textbf{58.8} \biggain{+28.1} \\
Communication Quality 
& \textbf{68.1} & 61.9 \bigdrop{-6.2}
& \textbf{55.5} & 54.2 \smalldrop{-1.3} \\
Context Awareness 
& 41.7 & \textbf{53.4} \biggain{+11.7}
& 4.0 & \textbf{40.5} \biggain{+36.5} \\
Instruction Following 
& \textbf{61.2} & 59.4 \smalldrop{-1.8}
& 45.8 & \textbf{46.5} \posdelta{+0.7} \\
\bottomrule
\end{tabular}%
}
\end{table*}

In this experiment, we compare DeCode with a zero-shot baseline built on the same underlying LLM, OpenAI o3. Both methods are evaluated on the full HealthBench dataset as well as its hard subset, with results summarized in Table~\ref{tab:main_results}.

A clear performance gap emerges between the full dataset and the hard subset. On the full set, the zero-shot baseline performs weakest on \textit{health data tasks}. Performance further degrades on the hard subset, where the baseline struggles across nearly all conversation themes; the highest score achieved is only $31.8\%$ under the \textit{global health} theme, highlighting the increased difficulty of this split.

In contrast, DeCode improves response quality across all conversation themes on the full set, with the exception of the \textit{complex responses} category. Further analysis suggests that DeCode occasionally generates overly detailed responses for relatively simple or straightforward queries. While this behavior can enrich informational content, it may negatively affect perceived \textit{communication quality}, contributing to the observed performance drop along this evaluation axis.

On the hard subset, DeCode yields substantial gains in overall performance. Notably, the lowest-scoring \textit{health data} theme improves to $35.6\%$, while all remaining themes exceed $40\%$. These results underscore the effectiveness of DeCode in enhancing both the \textit{content} and \textit{delivery} of medical question answering under more challenging evaluation conditions.

\subsection{Generalizability Across Backbone Models}
\begin{table*}[t]
\centering
\caption{\textbf{DeCode performance across diverse base LLMs.} Comparison between zero-shot (ZS) and DeCode-enhanced performance on the HealthBench hard subset across multiple base LLMs. Inline deltas in the DeCode columns are computed relative to the ZS baseline for the same model; deltas larger than 2 points are highlighted.}
\label{tab:model_results}
\resizebox{0.93 \textwidth}{!}{%
\renewcommand{\arraystretch}{1.0}
\begin{tabular}{l cc|cc|cc|cc}
\toprule
& \multicolumn{2}{c|}{\textbf{GPT-5.2}} & \multicolumn{2}{c|}{\textbf{OpenAI o3}} & \multicolumn{2}{c|}{\textbf{Claude 4.5}} & \multicolumn{2}{c}{\textbf{DeepSeek R1}} \\
\cmidrule(lr){2-3} \cmidrule(lr){4-5} \cmidrule(lr){6-7} \cmidrule(lr){8-9}
\textbf{Metric}~($\uparrow$, \%) & \textbf{ZS} & \textbf{DeCode} & \textbf{ZS} & \textbf{DeCode} & \textbf{ZS} & \textbf{DeCode} & \textbf{ZS} & \textbf{DeCode} \\
\midrule
\rowcolor{gray!10} \textbf{Overall Score} 
& 36.6 & \textbf{56.0} \biggain{+19.4}
& 28.4 & \textbf{49.8} \biggain{+21.4}
& 12.4 & \textbf{40.0} \biggain{+27.6}
& 14.8 & \textbf{25.7} \biggain{+10.9} \\
\midrule
\multicolumn{9}{l}{\textit{\textbf{Themes}}} \\
Emergency Ref. 
& 53.6 & \textbf{65.8} \biggain{+12.2}
& 27.0 & \textbf{59.1} \biggain{+32.1}
& 18.5 & \textbf{50.2} \biggain{+31.7}
& 19.9 & \textbf{33.0} \biggain{+13.1} \\
Context Seeking 
& 40.7 & \textbf{62.6} \biggain{+21.9}
& 30.0 & \textbf{58.3} \biggain{+28.3}
& 12.1 & \textbf{46.8} \biggain{+34.7}
& 15.0 & \textbf{32.0} \biggain{+17.0} \\
Global Health 
& 31.9 & \textbf{51.8} \biggain{+19.9}
& 31.8 & \textbf{49.8} \biggain{+18.0}
& 7.3 & \textbf{33.9} \biggain{+26.6}
& 15.6 & \textbf{22.3} \biggain{+6.7} \\
Health Data Tasks 
& 34.0 & \textbf{52.1} \biggain{+18.1}
& 17.0 & \textbf{35.6} \biggain{+18.6}
& 15.0 & \textbf{35.4} \biggain{+20.4}
& 3.1 & \textbf{22.0} \biggain{+18.9} \\
Communications 
& 33.8 & \textbf{55.7} \biggain{+21.9}
& 29.2 & \textbf{43.8} \biggain{+14.6}
& 16.6 & \textbf{39.8} \biggain{+23.2}
& 16.1 & \textbf{18.0} \posdelta{+1.9} \\
Hedging 
& 37.8 & \textbf{60.7} \biggain{+22.9}
& 30.9 & \textbf{54.6} \biggain{+23.7}
& 13.1 & \textbf{48.7} \biggain{+35.6}
& 18.3 & \textbf{34.1} \biggain{+15.8} \\
Complex Resp. 
& 35.1 & \textbf{44.0} \biggain{+8.9}
& 24.4 & \textbf{41.3} \biggain{+16.9}
& 15.2 & \textbf{27.5} \biggain{+12.3}
& 14.8 & \textbf{15.9} \posdelta{+1.1} \\
\midrule
\multicolumn{9}{l}{\textit{\textbf{Axes}}} \\
Accuracy 
& 49.4 & \textbf{62.6} \biggain{+13.2}
& 45.6 & \textbf{60.9} \biggain{+8.7}
& 28.6 & \textbf{49.5} \biggain{+20.9}
& 30.5 & \textbf{32.6} \biggain{+2.1} \\
Completeness 
& 24.9 & \textbf{56.6} \biggain{+31.7}
& 30.7 & \textbf{58.8} \biggain{+28.1}
& 3.7 & \textbf{44.1} \biggain{+40.4}
& 15.6 & \textbf{27.8} \biggain{+12.2} \\
Comm. Quality 
& \textbf{68.0} & 54.8 \bigdrop{-13.2}
& \textbf{55.5} & 54.2 \smalldrop{-1.3}
& \textbf{67.3} & 53.5 \bigdrop{-13.8}
& \textbf{60.9} & 58.4 \bigdrop{-2.5} \\
Cont. Awareness 
& 33.0 & \textbf{50.7} \biggain{+17.7}
& 4.0 & \textbf{40.5} \biggain{+36.5}
& 1.5 & \textbf{30.9} \biggain{+29.4}
& 0.0 & \textbf{19.1} \biggain{+19.1} \\
Inst. Following 
& \textbf{56.8} & 48.0 \bigdrop{-8.8}
& 45.8 & \textbf{46.5} \posdelta{+0.7}
& \textbf{45.7} & 43.6 \bigdrop{-2.1}
& \textbf{44.8} & 42.5 \bigdrop{-2.3} \\
\bottomrule
\end{tabular}%
}
\end{table*}

In this experiment, we examine the generalizability of DeCode across different base LLMs. A key advantage of the proposed framework is its model-agnostic design, which allows it to be applied to a wide range of base LLMs while consistently improving medical question answering performance. We evaluate DeCode on the hard subset using several leading LLMs from different providers, with results reported in Table~\ref{tab:model_results}.

Based on the zero-shot performance of the base LLMs, \textit{health data tasks} emerge as a particularly challenging category for GPT-5.2, OpenAI o3, and DeepSeek R1. In contrast, Claude-4.5 exhibits its weakest performance on \textit{global health} tasks. Across evaluation axes, \textit{context awareness} and \textit{completeness} are especially challenging: OpenAI o3, Claude-4.5, and DeepSeek R1 all record single-digit scores on these dimensions in certain cases, indicating systematic deficiencies in handling complex contextual and informational requirements.

Consistent with the observations in Section~\ref{results:main_results}, DeCode delivers substantial improvements over the corresponding zero-shot baselines across all tested models. Notably, Claude-4.5, which attains an initial overall score of $12.4\%$, improves to $40.0\%$ when integrated with DeCode. Similarly, the strongest baseline model, GPT-5.2, improves from $36.6\%$ to $56.0\%$. Importantly, in all experiments the underlying base LLM remains unchanged. By explicitly decoupling \textit{content} from \textit{delivery}, DeCode systematically enhances the performance of diverse base LLMs across nearly all medical QA scenarios. These results demonstrate that the benefits of DeCode are robust and largely LLM-agnostic, extending across a wide range of model architectures and providers.


\subsection{Comparison with LLM Frameworks}
\begin{table*}[!t]
\centering
\caption{\textbf{Comparison with leading LLM-based frameworks.} We evaluate MDAgents, KAMAC, and MuSeR, together with standard zero-shot prompting, a single-stage baseline, and DeCode. All methods use OpenAI~o3 on the HealthBench hard subset. Inline deltas are reported relative to the zero-shot baseline, with improvements greater than 2 points highlighted.}
\label{tab:sota_comparison_full}
\resizebox{0.97 \textwidth}{!}{%
\renewcommand{\arraystretch}{1.0}
\setlength{\tabcolsep}{3.5pt} 
\begin{tabular}{l c|c|c|c|c|c}
\toprule
\textbf{Metric}~($\uparrow$, \%) & 
\makebox[2cm][c]{\textbf{Zero-Shot}} & 
\makebox[2cm][c]{\textbf{Single Stage}} & 
\makebox[2cm][c]{\textbf{MDAgents}} & 
\makebox[2cm][c]{\textbf{KAMAC}} & 
\makebox[2cm][c]{\textbf{MuSeR}} & 
\makebox[2cm][c]{\textbf{DeCode}} \\
\midrule
\rowcolor{gray!15} \textbf{Overall Score}
& 28.4
& 39.1 \biggain{+10.7}
& 36.2 \biggain{+7.8}
& 27.4 \smalldrop{-1.0}
& 47.1 \biggain{+18.7}
& \textbf{49.8} \biggain{+21.4} \\
\midrule
\multicolumn{7}{l}{\textit{\textbf{Themes}}} \\
Emergency Referrals
& 27.0
& 53.4 \biggain{+26.4}
& 36.3 \biggain{+9.3}
& 33.1 \biggain{+6.1}
& 53.3 \biggain{+26.3}
& \textbf{59.1} \biggain{+32.1} \\
Context Seeking
& 30.0
& 44.8 \biggain{+14.8}
& 34.4 \biggain{+4.4}
& 27.4 \bigdrop{-2.6}
& \textbf{60.1} \biggain{+30.1}
& 58.3 \biggain{+28.3} \\
Global Health
& 31.8
& 38.7 \biggain{+6.9}
& 43.0 \biggain{+11.2}
& 30.2 \smalldrop{-1.6}
& 44.4 \biggain{+12.6}
& \textbf{49.8} \biggain{+18.0} \\
Health Data Tasks
& 17.0
& 24.0 \biggain{+7.0}
& 19.1 \biggain{+2.1}
& 7.7 \bigdrop{-9.3}
& \textbf{37.2} \biggain{+20.2}
& 35.6 \biggain{+18.6} \\
Communications
& 29.2
& 33.6 \biggain{+4.4}
& 38.0 \biggain{+8.8}
& 32.9 \biggain{+3.7}
& 40.4 \biggain{+11.2}
& \textbf{43.8} \biggain{+14.6} \\
Hedging
& 30.9
& 45.2 \biggain{+14.3}
& 39.2 \biggain{+8.3}
& 29.9 \smalldrop{-1.0}
& 51.2 \biggain{+20.3}
& \textbf{54.6} \biggain{+23.7} \\
Complex Responses
& 24.4
& 33.3 \biggain{+8.9}
& 31.6 \biggain{+7.2}
& 28.5 \biggain{+4.1}
& 37.7 \biggain{+13.3}
& \textbf{41.3} \biggain{+16.9} \\
\midrule
\multicolumn{7}{l}{\textit{\textbf{Axes}}} \\
Accuracy
& 45.6
& 43.6 \bigdrop{-2.0}
& 52.9 \biggain{+7.3}
& 43.8 \smalldrop{-1.8}
& 53.5 \biggain{+7.9}
& \textbf{54.3} \biggain{+8.7} \\
Completeness
& 30.7
& 44.5 \biggain{+13.8}
& 45.8 \biggain{+15.1}
& 36.1 \biggain{+5.4}
& 49.1 \biggain{+18.4}
& \textbf{58.8} \biggain{+28.1} \\
Communication Quality
& 55.5
& \textbf{59.9} \biggain{+4.4}
& 49.5 \bigdrop{-6.0}
& 46.7 \bigdrop{-8.8}
& 54.1 \smalldrop{-1.4}
& 54.2 \smalldrop{-1.3} \\
Context Awareness
& 4.0
& 29.9 \biggain{+25.9}
& 4.6 \posdelta{+0.6}
& 0.0 \bigdrop{-4.0}
& \textbf{46.0} \biggain{+42.0}
& 40.5 \biggain{+36.5} \\
Instruction Following
& 45.8
& 44.7 \smalldrop{-1.1}
& \textbf{53.5} \biggain{+7.7}
& 36.2 \bigdrop{-9.6}
& 46.2 \posdelta{+0.4}
& 46.5 \posdelta{+0.7} \\
\bottomrule
\end{tabular}%
}
\end{table*}
In this experiment, we evaluate representative LLM-based medical QA frameworks on the HealthBench hard subset using OpenAI~o3 as the base LLM. Specifically, we consider a Single Stage baseline, MDAgents~\cite{kim2024mdagents}, KAMAC~\cite{wu2025kamac}, and MuSeR~\cite{zhou2025muser}, with results summarized in Table~\ref{tab:sota_comparison_full}.

To isolate DeCode's structural benefits, we introduce a single-stage baseline that compresses our framework into a single prompt, forcing the LLM to internally process all steps before responding. Although the single-stage baseline improves over zero-shot prompting, it still lags behind DeCode by a significant margin. This gap highlights the importance of decoupling modules: requiring a single LLM call to simultaneously manage context extraction, clinical reasoning, and discourse planning exceeds what the model can reliably coordinate within one generation pass. By explicitly separating these tasks, DeCode systematically reduces this burden, achieving the highest overall performance among all evaluated methods.

Relative to the zero-shot baseline, MDAgents demonstrates consistent improvements across all conversation themes and four of the five evaluation axes. Notably, it achieves the strongest performance on \textit{instruction following} among all compared methods. This success stems from its \textit{complexity-driven} orchestration strategy: by estimating query difficulty upfront to form a fixed expert team, it ensures stable role assignment and coherent multi-agent collaboration. This static team composition appears particularly effective for instruction-heavy medical QA scenarios.

In contrast, KAMAC yields inconsistent gains over the zero-shot baseline and exhibits notable degradations in \textit{context awareness}, \textit{communication quality}, and \textit{instruction following}. Its \textit{knowledge-driven} strategy dynamically recruits specialists mid-generation to address missing domain knowledge. While this adaptive recruitment mechanism is intended to enhance coverage, our analysis suggests that introducing new experts mid-stream can disrupt conversational coherence. Specifically, the newly added agents often generate responses that overlap with existing contributions or shift the discussion focus, leading to redundancy and task-level confusion. These effects are amplified in longer, multi-round discussions, ultimately degrading response quality.

Among the compared LLM-based frameworks, MuSeR’s self-refinement approach yields the largest performance gains, particularly on \textit{context seeking} and \textit{health data tasks}, and achieves the strongest improvement in \textit{context awareness}. This is consistent with MuSeR’s design, where structured self-assessment guides targeted response refinement to recover missing contextual signals. Our results show that such self-refinement remains effective even when applied purely at inference time. However, MuSeR does not explicitly disentangle clinical content from discourse strategy, limiting its ability to independently optimize medical accuracy and communicative quality—an aspect directly addressed by DeCode.


Taken together, these results indicate that medical QA performance depends critically on how contextual information and communicative intent are structured during generation. While fixed-team orchestration and self-refinement improve context sensitivity, dynamic expert recruitment incurs overhead. Explicitly decoupling clinical content from discourse strategy provides a more stable architecture, enabling balanced optimization of medical accuracy and communication quality, as demonstrated by DeCode.




\subsection{Ablation Study}
\begin{table*}[t]
\centering
\caption{\textbf{Ablation study on the HealthBench hard subset.} DeCode denotes the complete framework. Each ablation independently removes a single component (Profiler, Formulator, or Strategist), while keeping the remaining modules unchanged. Values report absolute scores, with inline deltas indicating changes relative to DeCode. Deltas larger than 2 points are highlighted.}
\label{tab:ablation_full}
\resizebox{ 0.9\textwidth}{!}{
\renewcommand{\arraystretch}{0.92}

\begin{tabular}{l c|c|c|c}
\toprule
\textbf{Metric}~($\uparrow$, \%) 
& \textbf{DeCode} 
& \textbf{w/o Profiler} 
& \textbf{w/o Formulator} 
& \textbf{w/o Strategist} \\
\midrule
\rowcolor{gray!15} \textbf{Overall Score} 
& 49.8 
& 49.3 \smalldrop{-0.5} 
& 39.7 \bigdrop{-10.1} 
& 49.4 \smalldrop{-0.4} \\
\midrule
\multicolumn{5}{l}{\textit{\textbf{Themes}}} \\
Emergency Referrals 
& 59.1 
& 61.3 \biggain{+2.2} 
& 52.9 \bigdrop{-6.2} 
& 57.9 \smalldrop{-1.2} \\
Context Seeking 
& 58.3 
& 57.6 \smalldrop{-0.7} 
& 44.2 \bigdrop{-14.1} 
& 59.4 \posdelta{+1.1} \\
Global Health 
& 49.8 
& 50.7 \posdelta{+0.9} 
& 39.6 \bigdrop{-10.2} 
& 51.5 \posdelta{+1.7} \\
Health Data Tasks 
& 35.6 
& 34.2 \smalldrop{-1.4} 
& 32.1 \bigdrop{-3.5} 
& 40.0 \biggain{+4.4} \\
Communications 
& 43.8 
& 41.4 \bigdrop{-2.4} 
& 33.6 \bigdrop{-10.2} 
& 43.0 \smalldrop{-0.8} \\
Hedging 
& 54.6 
& 55.5 \posdelta{+0.9} 
& 43.9 \bigdrop{-10.7} 
& 54.1 \smalldrop{-0.5} \\
Complex Responses 
& 41.3 
& 35.4 \bigdrop{-5.9} 
& 30.8 \bigdrop{-10.5} 
& 30.3 \bigdrop{-11.0} \\
\midrule
\multicolumn{5}{l}{\textit{\textbf{Axes}}} \\
Accuracy 
& 54.3 
& 54.1 \smalldrop{-0.2} 
& 47.5 \bigdrop{-6.8} 
& 52.6 \smalldrop{-1.7} \\
Completeness 
& 58.8 
& 58.0 \smalldrop{-0.8} 
& 43.9 \bigdrop{-14.9} 
& 59.7 \posdelta{+0.9} \\
Communication Quality 
& 54.2 
& 54.4 \posdelta{+0.2} 
& 59.0 \biggain{+4.8} 
& 47.7 \bigdrop{-6.5} \\
Context Awareness 
& 40.5 
& 39.6 \smalldrop{-0.9} 
& 29.6 \bigdrop{-10.9} 
& 44.2 \biggain{+3.7} \\
Instruction Following 
& 46.5 
& 47.9 \posdelta{+1.4} 
& 49.2 \biggain{+2.7} 
& 44.2 \bigdrop{-2.3} \\
\bottomrule
\end{tabular}%
}
\end{table*}

We conduct an ablation study to assess the individual contribution of each component in the DeCode framework. Specifically, we independently remove the Profiler, Formulator, and Strategist modules and compare their performance against the full DeCode model on the HealthBench hard subset. The results are summarized in Table~\ref{tab:ablation_full}.

\paragraph{Impact of the Profiler}
The Profiler module is designed to extract the user’s background and underlying needs, enabling personalized response generation. Removing this component is therefore expected to reduce personalization in scenarios that require a deeper understanding of the user. As shown in Table~\ref{tab:ablation_full}, the absence of the Profiler leads to notable performance drops in \textit{ communications} and \textit{complex responses}. These degradations align with our expectations, as removing the Profiler limits the model’s ability to infer the appropriate level of detail and tailor responses to user-specific contexts.

\paragraph{Impact of the Formulator}
The Formulator module is responsible for identifying and structuring salient clinical indicators from the conversation history. Without this module, the model must rely on unstructured context, which can hinder coherent reasoning over clinical details. Consistent with this intuition, removing the Formulator results in substantial declines in \textit{completeness} and \textit{context awareness}, along with a modest reduction in \textit{accuracy}. These findings highlight the importance of the Formulator in organizing clinical information and ensuring that relevant conditions are explicitly addressed during medical question answering.

\paragraph{Impact of the Strategist}
The Strategist module governs response delivery by shaping tone, framing, and discourse strategy. Its removal primarily affects how information is communicated rather than what information is presented. As observed in our results, ablating the Strategist leads to a pronounced drop in \textit{communication quality} and a smaller but consistent decline in \textit{instruction following}. Both axes reflect how effectively responses engage with and adapt to user expectations. These results underscore the role of the Strategist in ensuring that medically relevant content is conveyed in an appropriate, user-receptive manner.

\section{Conclusion}
\label{sec:conclusion}
In this work, we introduce \textbf{Decoupling Content and Delivery (DeCode)}, a modular framework for contextualized medical question answering. DeCode adapts a base LLM into four specialized components---\textbf{Profiler, Formulator, Strategist, and Synthesizer}---that jointly structure response generation by explicitly separating medical content reasoning from discourse and delivery. This design enables the model to produce medically accurate responses while remaining sensitive to user context and communication needs.

Experiments on the OpenAI HealthBench benchmark demonstrate that DeCode consistently outperforms a zero-shot baseline and remains competitive with leading multi-agent frameworks across both full and hard evaluation settings. Moreover, evaluations across multiple base LLMs show that DeCode generalizes well across different model families and architectures, highlighting its model-agnostic nature.

Future work may explore mechanisms for caching and updating patient-specific information across multi-round interactions. Additionally, extending the DeCode paradigm beyond medical QA to other user-centered domains represents a promising direction. Taken together, these results suggest that DeCode provides a principled foundation for advancing contextualized medical QA.

\vfill

\section{Limitations}
Our evaluation is conducted on simulated patient--clinician conversations, which may not fully reflect the complexity, uncertainty, and risk profiles of real-world clinical settings. Although DeCode improves response quality without additional training, outputs generated by large language models may still contain errors or omissions and should not be used as a substitute for professional medical judgment. Validation and safeguards are necessary before deploying systems in clinical practice.

\bibliography{custom}

\clearpage
\appendix

\section{Cost \& Latency Analysis}

In this section, we analyze the cost and latency of the DeCode framework. In addition, we include the cost and latency of MDAgents~\cite{kim2024mdagents}, KAMAC~\cite{wu2025kamac}, and MuSeR~\cite{zhou2025muser}. To ensure fair comparison, we assume an output token rate of $70$ tokens/second to avoid latency fluctuations due to API re-routing. Furthermore, costs are estimated based on the official OpenAI o3 pricing model, $\$2.00$ per 1M input tokens and $\$8.00$ per 1M output tokens. 

\paragraph{DeCode Component Breakdown.} Table~\ref{tab:component_stats} details the token usage and latency of each module within our framework. By executing the Profiler (Background and User Needs) and the Formulator modules concurrently, we effectively mask the latency of the initial extraction steps. The final Synthesizer accounts for the largest portion of the overall latency ($20.83$\,s), as it is responsible for generating the comprehensive, patient-facing response. Overall, the parallelized architecture strictly bounds the total latency to approximately $45$ seconds per query.

\paragraph{Comparative Efficiency.} Table~\ref{tab:comparative_stats} highlights the substantial computational overhead commonly observed in complex multi-agent systems. Frameworks like KAMAC and MDAgents incur massive token overheads due to redundant context-sharing and lengthy debate loops. This results in prohibitively high costs and impractical latencies. In contrast, DeCode's decoupled, linear pipeline avoids the extensive communication overhead inherent in multi-agent frameworks. Consequently, DeCode achieves the lowest average cost ($\$0.037$ per query) and latency ($45.47$\,s) among all evaluated frameworks.



\begin{table*}[t]
    \centering
    \caption{\textbf{Token usage and latency breakdown of the DeCode framework.} We report average input tokens, output tokens, and estimated latency for each DeCode module. Latency is computed assuming a standardized throughput of 70 tokens/s. Since the Profiler and Formulator modules execute in parallel, their combined latency is attributed to the Formulator module.}
    \label{tab:component_stats}
    \resizebox{0.95\textwidth}{!}{%
        \begin{tabular}{lrrr}
            \toprule
            \textbf{Component} & \textbf{Avg. Input Tok.} & \textbf{Avg. Output Tok.} & \textbf{Avg. Latency (s)} \\
            \midrule
            Background (Profiler)        & 426.71 & 229.51 & 3.29 \\
            User Needs (Profiler)        & 360.71 & 213.31 & 3.04 \\
            Clinical Indicators (Formulator) & 515.71 & 876.54 & 12.53 \\
            Discourse Strategy (Strategist)  & 1,420.37 & 848.34 & 12.11 \\
            Response (Synthesizer)       & 1,502.16 & 1,457.63 & 20.83 \\
            \midrule
            \textbf{DeCode Total (parallelized)}      & \textbf{4,225.66} & \textbf{3,625.33} & \textbf{45.47} \\
            \bottomrule
        \end{tabular}%
    }
\end{table*}

\begin{table*}[t]
    \centering
    \caption{\textbf{Comparison of computational cost and latency across methods.} We report average input tokens, output tokens, estimated cost, and latency per sample. Latency is computed assuming a standardized generation throughput of 70 tokens/s. \textbf{Bold} denotes the best-performing method in each column (lowest cost or lowest latency).}
    \label{tab:comparative_stats}
    \resizebox{0.95\textwidth}{!}{%
        \begin{tabular}{lrrrr}
            \toprule
            \textbf{Method} & \textbf{Avg. Input Tok.} & \textbf{Avg. Output Tok.} & \textbf{Avg. Cost (\$)} & \textbf{Avg. Latency (s)} \\
            \midrule
            MDAgents & 44,444 & 4,909 & 0.128 & 70.13 \\
            KAMAC & 574,438.50 & 29,610.58 & 1.386 & 423.01 \\
            MuSeR & 6,205.93 & 2437.13  & 0.040 & 48.96 \\
            \textbf{DeCode (Ours)} & 4,225.66  & 3,625.33 & \textbf{0.037} & \textbf{45.47} \\
            \bottomrule
        \end{tabular}%
    }
\end{table*}

\section{Prompt Templates}
\label{sec:appendix_prompts}
The prompts for each module of \textbf{DeCode} are provided below. The Profiler module uses two independent prompts to extract the user background $\mathcal{B}$ and user needs $\mathcal{N}$. The prompt to extract the user background $\mathcal{B}$ and user needs $\mathcal{N}$ is provided in Figure \ref{fig:prompt_stage1_background} and Figure \ref{fig:prompt_stage1_need}. The Formulator prompt to extract the clinical indicators $\mathcal{C}$ is listed in Figure \ref{fig:prompt_stage2_content}. The Strategist prompt is given in Figure \ref{fig:prompt_stage3_strategy}. Finally, the Synthesizer prompt is presented in Figure \ref{fig:prompt_stage4_generation}.

\begin{figure*}[t]
\begin{promptbox}{User Background ($\mathcal{B}$)}
\ttfamily\small
\raggedright

You are a medical intake specialist. Analyze the following conversation and extract the user's background information.

\vspace{1em}
CONVERSATION:\\
\{conversation\_history\}

\vspace{1em}
Extract and infer the following information about the user (if available in the conversation):\\
- Age or age group\\
- Career/Occupation\\
- Financial Constraints (inferred from context)\\
- Living place/location\\
- Living situation (alone, with family, etc.)\\
- Any other relevant personal context

\vspace{1em}
IMPORTANT: Only include information that can be reasonably inferred from the conversation. Do NOT make up information.

\vspace{1em}
Respond in this EXACT format:

\vspace{1em}
AGE: [age or age group, or "Not specified"]\\
CAREER: [occupation, or "Not specified"]\\
ECONOMIC\_CONDITION: [economic status inferred from context, or "Not specified"]\\
LIVING\_PLACE: [location/region, or "Not specified"]\\
LIVING\_SITUATION: [living arrangement, or "Not specified"]\\
OTHER\_CONTEXT: [any other relevant information, or "None"]

\vspace{1em}
Be concise and factual. If information is not available, write "Not specified" or "None".

\end{promptbox}
\caption{Prompt template for the User Background extraction.}
\label{fig:prompt_stage1_background}
\end{figure*}

\begin{figure*}[t]
\begin{promptbox}{User Need ($\mathcal{N}$)}
\ttfamily\small
\raggedright 

You are analyzing a medical conversation to understand what the user needs.

\vspace{1em}
CONVERSATION:\\
\{conversation\_history\}

\vspace{1em}
Identify what the user explicitly asks for or clearly needs. Be conservative - only include needs that are:\\
1. Explicitly stated by the user\\
2. Clearly implied by the user's questions or concerns

\vspace{1em}
DO NOT include:\\
- Things the user might need but didn't mention\\
- General medical advice that wasn't requested\\
- Assumptions about what the user should want

\vspace{1em}
Respond in this EXACT format:

\vspace{1em}
NEEDS:\\
1. [First explicit need]\\
2. [Second explicit need]\\
3. [Third explicit need]\\
...

\vspace{1em}
If the user doesn't clearly state what they want, respond with:\\
NEEDS:\\
None specified

\vspace{1em}
Be strict and conservative.
\end{promptbox}
\caption{Prompt template for the User Need identification.}
\label{fig:prompt_stage1_need}

\begin{promptbox}{Clinical Indicators ($\mathcal{C}$)}
\ttfamily\small
\raggedright

You are a clinical safety and completeness planner.

\vspace{1em}
Your ONLY job is to identify the medically important content that MUST be covered\\
for this case to be safe, accurate, and reasonably complete. You are NOT deciding tone or style.\\
You are optimizing for clinical accuracy and completeness, not brevity.

\vspace{1em}
CONVERSATION:\\
\{conversation\_history\}

\vspace{1em}
Create a numbered list of key clinical content items that the final answer should try to cover, such as:\\
- Important symptom details or history that should be addressed or clarified\\
- Key possible causes or differentials (described in a cautious, non-diagnostic way)\\
- Red-flag or emergency warning signs that should be mentioned if relevant\\
- What the user can monitor or do at home (if appropriate)\\
- When and how urgently they should seek in-person care\\
- Any important limitations or uncertainties of online advice

\vspace{1em}
Rules:\\
- Focus on clinical content ONLY (WHAT to cover), not HOW to phrase it.\\
- Err on the side of including any clinically important point that might affect safety.\\
- Each item should be 1--2 sentences max.\\
- Avoid repeating the same content in multiple items.\\
- Do not invent new symptoms; only build on what is in the conversation.\\
- It is acceptable to mention reasonable possible causes or scenarios even if the user did not use those exact words, as long as they logically follow from the described symptoms.

\vspace{1em}
Respond in this EXACT format:\\
1. [Clinical content item]\\
2. [Clinical content item]\\
3. [Clinical content item]\\
...

\end{promptbox}
\caption{Prompt template for the Formulator module ($\mathcal{M}_{form}$).}
\label{fig:prompt_stage2_content}
\end{figure*}

\begin{figure*}[t]
\begin{promptbox}{Discourse Strategy ($\mathcal{S}$)}
\ttfamily\small
\raggedright
\linespread{0.98}\selectfont

You are a response-strategy planner for a medical assistant.

\vspace{1em} 
You receive:\\
- The original conversation\\
- A brief user background profile\\
- A list of what the user clearly needs\\
- A clinical content checklist (what should be covered for safety/completeness)

\vspace{1em} 
Your job is to design HOW the assistant should answer for THIS user: what to prioritize,
how deep to go, what style and structure to use, and what to avoid.

\vspace{1em} 
CONVERSATION:\\
\{conversation\_history\}

\vspace{1em} 
USER BACKGROUND PROFILE:\\
\{user\_profile\}

\vspace{1em} 
USER NEEDS (what the user clearly wants):\\
\{needs\_formatted\}

\vspace{1em} 
CLINICAL CONTENT CHECKLIST (what should be covered):\\
\{content\_formatted\}

\vspace{1em} 
Pay particular attention to:\\
- Whether the user's needs are clearly stated or vague/unspecified.\\
- Whether there is sufficient information available for a safe medical assessment.\\
- When needs or information are unclear, the plan should usually include a brief strategy for clarifying key gaps (e.g., 1--2 focused questions), while still guiding the assistant to give the best possible provisional answer based only on what is already known.

\vspace{1em} 
IMPORTANT:\\
- The assistant MUST still give concrete, practical, medically useful information even when information is incomplete. Use conditional language (e.g., "If X..., then Y...") rather than refusing to say anything.\\
- Do NOT tell the assistant to avoid discussing possible causes or next steps entirely.\\
- Clarification questions should be few (0--2 of the most important ones) and should not dominate the answer.

\vspace{1em} 
Design a plan with TWO sections:

\vspace{1em} 
1. WHAT TO DO/COVER (TO DO):\\
   - How the assistant should prioritize and present the content for THIS user.\\
   - What level of technical detail is appropriate for this user.\\
   - Whether to keep the answer short vs. more detailed.\\
   - Whether to explicitly ask clarification questions (0--2 key questions only), and if so, in what style and at what point (usually after giving main guidance).\\
   - Which content items from the checklist are highest priority to cover explicitly.\\
   - How to adapt the response to the user's apparent role, location, and constraints.

\vspace{1em} 
2. WHAT NOT TO DO/COVER (NOT TO DO):\\
   - Things that would likely confuse, overwhelm, or frustrate THIS user.\\
   - Styles to avoid (e.g., too technical, too casual, too vague, overly long).\\
   - Types of content to avoid (e.g., extremely long, low-yield lists of differential diagnoses; strong reassurance when red flags are possible; rigid instructions when access is limited).\\
   - Any ways of answering that would clearly conflict with the user's instructions.

\vspace{1em} 
You are NOT writing the final medical answer. You are only writing the plan.

\vspace{1em} 
Respond in this EXACT format:

\vspace{1em} 
TO DO:\\
1. [Response strategy / priority tailored to user]\\
2. [Another response strategy / priority]\\
3. [Continue as needed]

\vspace{1em} 
NOT TO DO:\\
1. [Specific thing to avoid for this user]\\
2. [Another thing to avoid]\\
3. [Continue as needed]

\end{promptbox}
\caption{Prompt template for the Strategist module ($\mathcal{M}_{strat}$).}
\label{fig:prompt_stage3_strategy}
\end{figure*}

\begin{figure*}[t]
\begin{promptbox}{Controlled Generation ($R$)}
\ttfamily\small
\raggedright

You are an experienced medical professional providing personalized advice.

\vspace{1em}
Your highest priorities are:\\
1) Clinical accuracy and completeness of the information you provide.\\
2) Clear, practical guidance for the user.\\
3) Safe and appropriate communication.

\vspace{1em}
ORIGINAL CONVERSATION:\\
\{conversation\_history\}

\vspace{1em}
PRESENTATION GUIDELINES (HOW TO ANSWER):

\vspace{1em}
TO DO:\\
\{to\_do\_formatted\}

\vspace{1em}
NOT TO DO:\\
\{not\_to\_do\_formatted\}

\vspace{1em}
CONTENT CHECKLIST (WHAT YOU MUST COVER CLINICALLY):\\
\{content\_formatted\}

\vspace{1em}
Your task:\\
1. Cover ALL items in the CONTENT CHECKLIST as clearly and concretely as possible. Aim for at least one explicit sentence or short paragraph addressing each item.\\
2. Follow the TO DO / NOT TO DO guidelines for how to present the information in a way that fits THIS user's background and needs.\\
3. Be explicit about uncertainty and information gaps, but still give the BEST POSSIBLE DIRECT ANSWER based only on the conversation.\\
   - Use conditional language (e.g., "If X..., then Y...") rather than refusing to answer.\\
4. You may ask up to 1--2 of the most important clarification questions, but they should be placed near the end and should NOT replace giving guidance.\\
5. Keep the response user-centered and practical, and explain what the user can do next (e.g., monitor, self-care, when/where to seek in-person care).\\
6. End with a brief reminder that this information does not replace an in-person medical evaluation and that the user should seek care if they are worried or if concerning symptoms arise.

\vspace{1em}
Provide your response:
\end{promptbox}
\caption{Prompt template for the Synthesizer module ($\mathcal{M}_{syn}$).}
\label{fig:prompt_stage4_generation}
\end{figure*}

\section{Case Study Illustration}
\label{sec:appendix_case_study}

To provide a qualitative assessment of \textbf{DeCode}, we present a representative case directly from the HealthBench dataset.

\begin{table}[H] 
    \centering
    \small 
    \setstretch{1.15}
    \begin{tabularx}{\columnwidth}{@{} >{\raggedright\arraybackslash}X @{}} 
        \toprule
        \rowcolor{gray!10} \textbf{User Query (HealthBench Hard Subset)} \\
        \midrule
        \textit{``feeling depressed for the last few months''} \\
        \midrule
        
        \textbf{Baseline: Zero-Shot (o3) | Score: 0.259} \\
        \rowcolor{red!5} 
        I'm really sorry you've been feeling this way. Depression can be exhausting and isolating... 
        \newline
        1. \textbf{Talk to a professional:} \textit{[Generic advice omitted]}
        \newline
        2. \textbf{Reach out for support:} \textit{[Generic advice omitted]}
        \newline
        3. \textbf{Everyday self-care strategies:} \textit{[Generic advice omitted]}
        \newline
        4. \textbf{Keep track of feelings:} \textit{[Generic advice omitted]}
        \newline
        \textcolor{red}{\textbf{5. Crisis resources: If you ever feel like you might act on thoughts of harming yourself, call emergency services....}} \\
        \midrule
        
        \textbf{Ours: DeCode (o3) | Score: 0.907} \\
        \rowcolor{blue!5} 
        I'm really sorry you've been feeling depressed these past few months. You're not alone, and help is available. Because safety is the first priority, can I ask directly: 
        \newline
        \textcolor{blue}{\textbf{• Have you had any thoughts of ending your life or harming yourself?}} 
        \newline
        If you feel you might act on these thoughts, please call 911 right now...
        \newline
        To understand what you're experiencing and guide next steps, could you tell me a little more?
        \newline
        1. \textbf{Symptoms \& impact:} When did the low mood start? How is it affecting work, school, or self-care?
        \newline
        2. \textbf{Safety \& red flags:} Any thoughts of harming others? Have you noticed hearing or seeing things others don't?
        \newline
        3. \textbf{Bipolar screening:} Have you ever had periods of unusually high energy (mania)?
        \newline
        \textbf{Practical Next Steps:} Schedule a visit with a primary-care doctor. Evidence-based talk therapies (CBT) and medications may be appropriate. Keep a regular routine and stay connected. \\
        \bottomrule
    \end{tabularx}
    \caption{\textbf{Qualitative comparison on a mental health query.} DeCode (blue) prioritizes emergency triage, whereas the baseline (red) provides passive advice.}
    \label{tab:qualitative_comparison}
\end{table}

\paragraph{Qualitative Comparison.} 
According to the HealthBench grading results, the zero-shot baseline incurred a severe penalty ($-10$ points) for failing to assess immediate suicide risk, defaulting to passive advice. In contrast, DeCode earned full points ($+10$ points) for context awareness by explicitly screening for suicidal ideation and red flags. While both models received a verbosity penalty ($-5$ points) for providing over six suggestions, DeCode intentionally trades conversational brevity for comprehensive clinical safety and proper triage, yielding a vastly superior overall score ($0.907$ vs. $0.259$).

\end{document}